# Envelope Imbalance Learning Algorithm based on Multilayer Fuzzy C-means Clustering and Minimum Interlayer discrepancy

Fan Li, Xiaoheng Zhang, Pin Wang, Yongming Li*

School of Microelectronics and Communication Engineering, Chongqing University, Chongqing, China 400044

* Corresponding author: yongmingli@cqu.edu.cn; lymcentor924924@gmail.com

## Abstract

Imbalanced learning is important and challenging since the problem of the classification of imbalanced datasets is prevalent in machine learning and data mining fields. Sampling approaches are proposed to address this issue, and cluster-based oversampling methods have shown great potential as they aim to simultaneously tackle between-class and within-class imbalance issues. However, all existing clustering methods are based on a one-time approach. Due to the lack of a priori knowledge, improper setting of the number of clusters often exists, which leads to poor clustering performance. Besides, the existing methods are likely to generate noisy instances. To solve these problems, this paper proposes a deep instance envelope network-based imbalanced learning algorithm with the multilayer fuzzy c-means (MlFCM) and a minimum interlayer discrepancy mechanism based on the maximum mean discrepancy (MIDMD). This algorithm can guarantee high quality balanced instances using a deep instance envelope network in the absence of prior knowledge. First, the MlFCM is designed for the original minority class instances to obtain deep instances and increase the diversity of instances. Then, the MIDMD is proposed to avoid the generation of noisy instances and maintain the consistency of the interlayers of instances. Next, the multilayer FCM and minimum interlayer discrepancy mechanism are combined to construct a deep instance envelope network – the MlFC&IDMD. Finally, an imbalance learning algorithm is proposed based on the MlFC&IDMD. In the experimental section, thirty-three popular public datasets are used for verification, and over ten representative algorithms are used for comparison. The experimental results show that the proposed approach significantly outperforms other popular methods.

## 1. Introduction

In the data mining and machine learning research, a potential severe challenge is how to handle "imbalanced categories". One category may consist of a large number of examples while the other may have only a few. This is known as the class imbalance problem. The class imbalance problem has the following characteristics [1]: class overlapping, small sample size and small disjuncts. This causes a high degree of complications during the classification stage. When class imbalance is encountered, the overall classification accuracy of standard learning algorithms focuses too much on the majority class and degrades the classification performance of instances from the minority class, and classifiers may treat some of the data points in the minority class as outliers, which produces a very high misclassification error rate. Therefore, the primary class of interest in the data mining task is usually the minority (or rare) class, and it is necessary to improve the recognition accuracy of the minority class instances [2]. As the size of datasets become increasingly larger in numerous real-world applications, such as anomaly detection [3], medical diagnosis [4], target identification [5], and business analytics [6], the consequences of class imbalance problems become greater.

Existing approaches aimed at solving the class imbalance problem can be mainly categorized as follows: (1) Data level approaches: These approaches aim to rebalance the class prior distribution during preprocessing. These solutions are attractive because the only change needed is in the training data rather than the learning algorithm. Typical approaches include undersampling the majority class, oversampling the minority class and combining both [7]. (2) Algorithm level approaches: These approaches modify existing algorithms to place more emphasis on minority class instances. One typical strategy is cost-sensitive learning [8], which assigns different costs to different classes, e.g., higher penalties for minority class instances. However, it is often difficult to optimize cost matrices or relations. (3) Ensemble approaches: These approaches combine data level and algorithm level approaches to rebalance the data [9-11]. Ensemble-based classifiers, which are achieved by combining different resampling strategies (e.g., SMOTEBoost (Synthetic Minority Oversampling Technique with AdaBoost) [12], RUSBoost (random under-sampling with AdaBoost) [13], overbagging [14], and underbagging[15]), have been shown to be able to solve the imbalanced dataset problem. (4) Feature selection approaches: These approaches select a subset of all features that are more suitable for class imbalanced problems and reflect the characteristics of class imbalance to build a classification model so that the classifier can achieve better performance [16].

Data-level approaches are widely used due to their simple understanding and realization. Among the data-level approaches, random sampling including random undersampling and random oversampling is the easiest. However, random undersampling may lead to information loss, and random oversampling often leads to overfitting [17]. To address the overfitting problem, the SMOTE technique was introduced [18]. This method can avoid overfitting by generating new minority class instances through linear interpolation between K adjacent points rather than just duplicating existing instances. However, this approach may produce noisy instances due to its sensitivity to the value of K. Numerous extensions of SMOTE have been proposed. For instance, safety-level-SMOTE [19] calculates a "safety level" value for each minority instance and then generates synthetic instances that are closer to the maximum safety level, which is defined as the number of other minority instances in the nearest neighbors. Safety-level-SMOTE can lead to overfitting because synthetic instances are forced to be generated far from the decision boundary. Borderline-SMOTE [20] determines the borderline between two categories and oversamples only minority instances on the borderline. ADASYN [21] assigns weights to minority instances so that those with more majority instances in their neighborhoods have a higher probability of being oversampled. However, Borderline-SMOTE and ADASYN did not find all minority instances close to the decision boundary [22]. In general, it is necessary to find difficult-to-learn instances for oversampling as they contain important classification information. These instances are usually close to the decision boundary [23]. In general, there are two problems to be addressed. The first problem is the within-class imbalance problem, and the second problem is the noisy instances problem.

These algorithms aim to solve the between-class imbalance problem while ignoring the within-class imbalance problem, but SMOTE-based oversampling can lead to the generation of noisy instances because it does not guarantee that the generated instance distribution is closer to the original instance distribution [24-26]. To address the within-class imbalance problem, clustering algorithms are adopted. In clustering-based methods, first, the dataset is divided into several smaller subclusters, and then sampling methods are used in these subclusters to maintain class balance. Clustering algorithms can effectively solve the within-class imbalance problem.

Clustering-based methods can be divided into cluster-based undersampling and cluster-based oversampling. The cluster-based undersampling method [27] first decomposes the entire dataset into several subclusters and then removes some majority instances from each subcluster based on the imbalance ratio. Pattaramon et al presented an overlap-based undersampling approach to enhance the visibility of minority class instances in overlapping regions. This is achieved by identifying and eliminating negative instances in overlapping regions through soft clustering FCM and an elimination threshold [28]. Lin et al [29] developed a clustering-based undersampling method, which clusters only the majority classes and then uses the cluster centers or their nearest neighbors to represent the majority classes to maintain a balance with the minority classes. In addition, Ofek et al proposed an undersampling method based on fast clustering for solving the binary class imbalance problem. In this method, first, minority instances are clustered, and then a similar number of majority instances are selected to train a specific classifier [30].

Clustering-based oversampling methods are also widely used. Jo and Japcowicz proposed a method in which this method first clusters each class individually and then oversamples all subclusters of the same class to ensure that they are equivalent in size [31]. Cluster-SMOTE [32] first divides the minority class into several subclusters using the K-means algorithm and then oversamples each subcluster using SMOTE. In [33], the minority classes are clustered into several arbitrarily shaped subclusters, and then synthetic instances are generated between the minority instances and their corresponding pseudocenters of subclusters. Iman Nekooeimehr et al [34] proposed a new oversampling method, adaptive semiunsupervised weighted oversampling (A-SUWO). The method uses a semiunsupervised hierarchical clustering method to cluster the minority instances, and the minority instances are oversampled based on the Euclidean distance between the minority instances and the majority class. Xinmin Tao et al [35] proposed a novel adaptive weighted oversampling for imbalanced datasets based on density peak clustering with heuristic filtering (ADPCHFO). The proposed approach replaces traditional K-means clustering with modified density-peak clustering for minority instance clustering as it is able to accurately identify subclusters with different sizes and densities. Georgios Douzas et al [36] presented a new oversampling method, Self-Organizing Map-based Oversampling (SOMO). This method uses a self-organizing map to transform the input data into a two-dimensional space and applies SMOTE within clusters found in the lower dimensional space, as well as between neighboring clusters. Douzas et al [37] proposed a heuristic oversampling method based on the K-means and SMOTE (K-means SMOTE). This method uses the K-means to cluster the entire dataset into a number of subclusters and then uses SMOTE to oversample each subcluster. In general, undersampling approaches are less efficient than oversampling approaches as removing majority instances may result in the loss of important information in the dataset, especially for relatively small datasets.

Compared to cluster-based undersampling methods, clustering-based oversampling methods avoid the risk of missing instance information. However, it is worth noting that most of the current clustering-based oversampling algorithms are based on all instances or a single category of instances. These methods involve single-layer clustering and ignore the diversity of information between instances. It is difficult to explore more instance information using these methods. Conversely, single-layer clustering relies heavily on a priori knowledge, which limits its application, especially to complex datasets. A deep neural network can extract high-quality features by multiple layer transformations. Similarly, multiple layers of clustering can be used for sampling with multiple iterations. During times of clustering, the diversity of the interlayer can be smoothed out, and initialization settings of the number of clusters become less important. Therefore, it is necessary to consider multilayer clustering. Notably, in recent years, multilayer clustering algorithms have been proposed [38-39]. For example, R. Al-Hmouz et al proposed Hierarchical Granular Clustering (HGC) algorithm, and this algorithm can obtain global and local clusters and significantly improved the quality of clustering centroid [38]. These studies show that multilayer clustering helps to mine the structural information of instances and improve the clustering performance, but its purpose is still clustering rather than new instances generation for classification. Therefore, the number of clustering centroid is close to the final number of instances categories rather than instances, and these clustering methods also do not consider interlayer instances distribution and so on.

Regarding the second problem (noisy instances problem), it is necessary to consider eliminating the distributional discrepancy of the instances before and after clustering. To reduce the instance distributional discrepancy, the Maximum Mean Discrepancy (MMD) is adopted. The MMD has been widely used in domain adaptation [40] to measure the discrepancy of distributions and can be applied to impose constraints on the training models and optimized by updating the training parameters [41]. Therefore, optimizing the MMD can effectively reduce the discrepancy of the instance distribution and can be used to solve the noisy instances problem.

Existing clustering algorithms are the K-means, density peak clustering, FCM, SOM and others. Among these methods, the FCM, which was proposed by Bezek, is one of the most common soft clustering algorithms [42]. Soft clustering algorithms allow each instance to be a member of many clusters. The likelihood of belonging to a cluster is denoted as the degree of affiliation which has a value between 0 and 1; and the certain class to which data belong is determined by the affiliation function. The affiliation function makes the point not directly affiliated with a single clustering center, which results in the instances being more diverse and richer in information. Hence, FCM is chosen as the clustering algorithm for this paper. Based on the ideas above, a deep instance envelope network-based sampling fusion method called the MlFC&IDMD is proposed by combining the MlFCM and MIDMD. First, the multilayer FCM for the original minority class instances (MlFCM) is designed to obtain deep instances and increase the diversity of instances. Then, the MIDMD mechanism is proposed to avoid the generation of noisy instances. Finally, the multilayer

FCM and MIDMD mechanisms are combined. The generated deep instances make the two categories of instances in the original dataset reach a balance.

It should be emphasized that the MlFC&IDMD network proposed in this paper is fundamentally different from existing multilayer clustering methods. The main differences are: 1) The former is designed to generate new instances through MlFCM, while the latter is designed to gradually determine the clustering centroids (clustering prototypes) through hierarchical FCM. 2) Like deep stack autoencoders, unsupervised deep learning needs to maintain the consistency of interlayer of features, so the algorithm in this paper designs MIDMD to maintain the distribution consistency of the interlayers of instances; however, the purpose of existing multilayer clustering method is for clustering, so it does not consider this point. 3) As the purpose of existing multilayer clustering methods is to cluster and the number of clusters is much less than the number of instances, so its number of layers is less. However, the number of instance layers of the MlFC&IDMD network proposed in this paper can be relatively larger (in this paper, the number can reach 9). 4) Since the MlFC&IDMD network proposed in this paper is used to generate new instances, the number of clusters is much more than the number of clustered instances; however, the number of clusters of the multilayer clustering methods is much less than the number of instances and gradually approaches to reach the number of categories. 5) For existing multilayer clustering, only the highest level (upper level) is used for clustering, while the MlFC&IDMD network proposed in this paper integrates the clustering centroids (clustering instances) of each layer (level). 6) The MlFC&IDMD network proposed is only a part of this paper's algorithm, and it is used for minority classes to generate new instances through multilayer clustering. In addition, the algorithm in this paper also includes fusion of multilayer instances to obtain new minority class instances, recombination with majority class instances, classifier and so on. In general, existing multilayer clustering is effective in approximating the number of categories asymptotically by hierarchical clustering, which is beneficial for mining information about the cluster structure, but its final purpose is still clustering. Although this proposed method includes multilayer clustering, the purpose of the method is to generate new instances for classification rather than clustering. Besides, this proposed method includes other parts, including interlayer distribution consistency and multilayer new instances fusion, fusion of instances of different class, and so on.

## 2. Method

The proposed imbalance learning algorithm based on the MlFC&IDMD is related to the data-level oversampling algorithm. This algorithm uses MlFCM clustering to generate multilayer deep instances to enrich the original minority instance information and uses the MIDMD to make the distribution of instances before and after clustering consistent with each other to avoid the generation of noisy instances. That is, the new instances obtained by the MlFC&IDMD can not only enrich the information of original instances and make the instances more diverse but also guarantee that the new instances are consistent with the distribution of instances before clustering. The key notations are listed in Table 1, and Table 2 shows the related terminology of this paper.

Table 1. Key notations

| Notation | Definition |
|---|---|
| $u_{ik}$ | The degree of membership of $x_k$ to the $i$th cluster |
| $U = (u_{ik})_{c \times n}$ | The degree of membership matrix |
| $m$ | The fuzzification coefficient |
| $V^L$ | The $L$th-layer clustering prototypes |
| $C_L$ | The $L$th-layer number of clusters |
| $w$ | The number of iterations |
| $\Gamma_{MlFCM}$ | The MlFCM transformation operator |
| $1/t$ | The ratio of the number of instances generated in the current layer to the previous layer |
| $Maj$ | The number of majority instances |
| $Min$ | The number of minority instances |

Table 2. Related terminology used in this paper

| Related terminology | Definition |
|---|---|
| MlFCM | Multilayer Fuzzy C-Means Clustering |
| MIDMD | Minimum interlayer discrepancy mechanism based on maximum mean discrepancy |
| RKHS | Reproduced kernel Hilbert space |
| MlFC&IDMD | The deep instance envelope network combining MlFCM and MIDMD |
| FCM | Fuzzy C-Means Clustering |
| SlFCM | Single layer Fuzzy C-Means Clustering |
| MMD | Maximum mean discrepancy |
| MlFC&IDMD&IL | The proposed imbalance learning algorithm based on MlFC&IDMD |
| IR | Imbalance ratio |

## 2.1 Multilayer Fuzzy C-Mean Clustering (MlFCM) mechanism

Suppose a dataset of $n$ data points $X = \{x_1, x_2, ... x_n\} \in R^{n \times d}$ will be grouped into $C$ clusters $G_1, G_2, ..., G_c$, and the corresponding prototypes of clusters are denoted as $V = \{v_1, v_2, ..., v_c\} \in R^{c \times d}$. Apparently, the clusters can be considered envelopes of the data points. With fuzzy C-means clustering (FCM), the handling of the original instances can be transformed to the handling of the envelopes of the original instances, and the envelopes can be viewed as the high-level description of the original instances. The objective function of the FCM can be written as

$$\min J(U, V) = \sum_{i=1}^{c} \sum_{k=1}^{n} u_{ik}^m d_{ik}^2, \quad \text{s.t.} \sum_{i=1}^{c} u_{ik} = 1 \qquad (1)$$

where $u_{ik}$ is the degree of membership of $x_k$ to the $i$th cluster, and $U = (u_{ik})_{c \times n}$. $d_{ik} = x_k - v_i$ represents the Euclidean distance between $x_k$ and cluster center $v_i$. $m>1$ is called the fuzzification coefficient and is usually set to 2. The FCM is highly likely to produce a local minimum through the iterative formulas of the partition matrix and the prototypes

$$u_{ik} = \frac{1}{\sum_{j=1}^{c}\left[\frac{d_{ik}}{d_{jk}}\right]^{\frac{2}{m-1}}}, \quad v_i = \frac{\sum_{k=1}^{n}(u_{ik})^m x_k}{\sum_{k=1}^{n}(u_{ik})^m} \qquad (2)$$

Multilayer clustering is implemented based on single layer FCM (SlFCM) module. The original dataset $X^1 \in R^{n \times d}$ can be transformed into $V^1 = \{v_1, v_2, ..., v_{c_1}\} \in R^{c_1 \times d}$ by Eq. (2), a new dataset $X^2 = \{X^1, V^1\} \in R^{(n+c_1) \times d}$ can be grouped into $C_2$ clusters, and the new prototypes of clusters $V^2 = \{v_1, v_2, ..., v_{c_2}\} \in R^{c_2 \times d}$ can be obtained. If L-layer clustering is performed using the FCM, new instance points $V^L = \{v_1, v_2, ..., v_{c_L}\} \in R^{c_L \times d}$ will be obtained.

$$v_{i^L} = \frac{\sum_{k=1}^{n}(u_{i^L k})^m x_k + \sum_{i^1=1}^{C_1}(u_{i^L i^1})^m v_{i^1} + ... + \sum_{i^{L-1}=1}^{C_{L-1}}(u_{i^L i^{L-1}})^m v_{i^{L-1}}}{\sum_{k=1}^{n}(u_{i^L k})^m + \sum_{i^1=1}^{C_1}(u_{i^L i^1})^m + ... + \sum_{i^{L-1}=1}^{C_{L-1}}(u_{i^L i^{L-1}})^m}$$

$$i^L = 1, 2, ..., c_L \qquad (3)$$

where $C_1, ..., C_{L-1}$ are the number of clusters in the former $L$-1 layer. Through Eq. (3), hierarchical instance transformation is implemented by the FCM, so this method is called multilayer fuzzy C-means clustering (MlFCM). The overall scheme is portrayed in Figure 1. Figure 1(a) denotes a SlFCM module, and the instances can be transformed by the SlFCM module to obtain new instances. Figure 1(b) denotes Sample envelope based on FCM. Some relevant samples are combined into an envelope and transformed to a new instance (envelope instance) by FCM. Figure 1(c) denotes multilayer FCM mechanism based on SlFCM modules and the new instances for each layer can be obtained by instances transformation. The new instance is a set of original instances, so it is called 'envelope instance'. The pseudocode description for the MlFCM is also shown as follows.

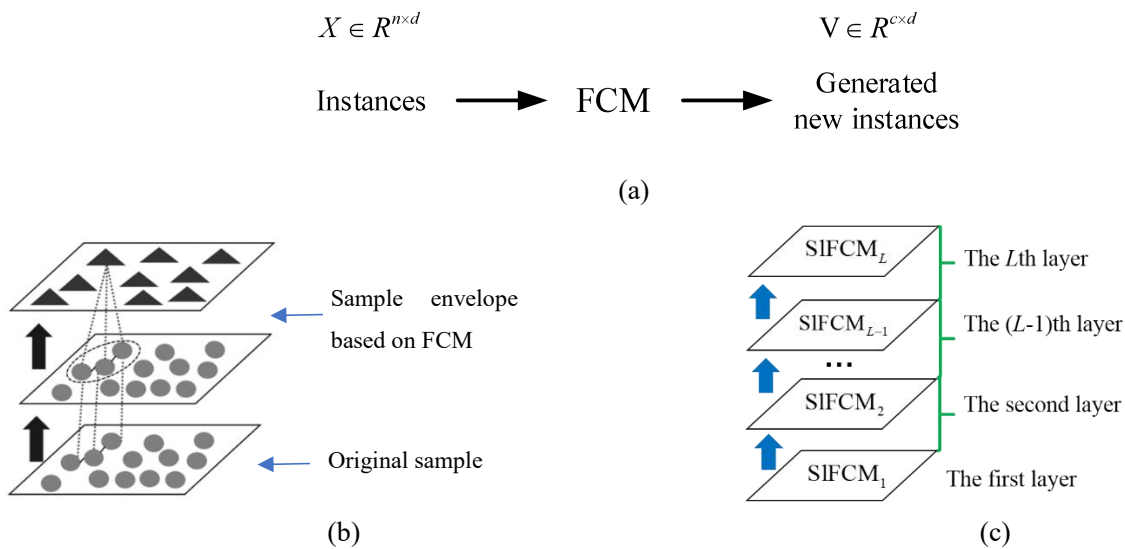

(a)

(b)

(c)

Fig. 1. The overall scheme of MlFCM:(a) SlFCM module; (b) Sample envelope based on FCM; (c) multilayer FCM mechanism based on SlFCM modules

---

*Algorithm 1: Multilayer Fuzzy C-Means Clustering (MlFCM)*

---

**Input:** Dataset $X$, the number of clusters per layer $C_1,..,C_L$, parameter $m$, iteration number $w$ and threshold $\varepsilon$.

**Output:** Generated new instances $V^s = \{V^1,...,V^L\}$

**Procedure:**

1. Initialize number of layers of clustering $L$;
2. For $l = 1:L$
3.     Initialize the degree of membership matrix $U$;
4.     $w \leftarrow 1$;
5.     **repeat**
6.         Update new instances $V^{l(w)} = \{v_1, v_2, ..., v_{c_l}\}$ via (3);
7.         Update degree of membership matrix $U^{(w)}$ via (2);
8.         $w \leftarrow w+1$;
9.     **until** $\left| J(U,V)^{(w+1)} - J(U,V)^{(w)} \right| < \varepsilon$;
10.    **return** $V^l$;
11. End
12. Return generated new instances set $V^s$.

---

## 2.2 Minimum interlayer discrepancy mechanism based on maximum mean discrepancy (MIDMD)

To reduce the distribution discrepancy, we adopt the empirical maximum mean discrepancy (MMD) as a nonparametric metric to measure the distribution discrepancy between the layers of the MlFCM. Previous studies have demonstrated that the MMD possesses more robust and efficient computation and optimization [43-44]. Therefore, the MMD is introduced into the MlFCM to reduce the distributional discrepancy of interlayer instances. The algorithm here is the minimum interlayer discrepancy mechanism based on the maximum mean discrepancy (MIDMD) and is described as follows.

Let $X = \{x_k\}_{k=1}^n$ and $V^1 = \{v_i\}_{i=1}^{c_1}$ represent the original dataset and the first layer of instances, respectively; and $\Gamma_{MlFCM}$ denotes the MlFCM transformation operator. The MMD between $X$ and $V^1$ is estimated as follows [45]:

$$D_{\mathrm{H}}^{2}\left(X,\mathrm{V}^{1}\right)=D_{\mathrm{H}}^{2}\left(X,\Gamma_{MlFCM}\left(X\right)\right)=\left\|\frac{1}{n}\sum_{k=1}^{n}f\left(x_{k}\right)-\frac{1}{c_{1}}\sum_{i=1}^{c_{1}}f\left(\Gamma_{MlFCM}\left(x_{k}\right)\right)\right\|_{\mathrm{H}}^{2} \quad (4)$$

If the reproduced kernel Hilbert space (RKHS) is constructed by using the characteristic kernels, Eq. (4) can be expanded as

$$D_{\mathrm{H}}^{2}\left(X,\mathrm{V}^{1}\right)=\left\|\frac{1}{n^{2}}\sum_{k=1}^{n}\sum_{k'=1}^{n}k\left(x_{k},x_{k'}\right)-\frac{2}{nc_{1}}\sum_{k=1}^{n}\sum_{i=1}^{c_{1}}k\left(x_{k},v_{i}\right)+\frac{1}{c_{1}^{2}}\sum_{i=1}^{c_{1}}\sum_{i'=1}^{c_{1}}k\left(v_{i},v_{i'}\right)\right\| \quad (5)$$

where

$$v_{i}=\Gamma_{MlFCM}(x_{k}) \quad (6)$$

Therefore, the distributional discrepancy between the instance sets $X$ and $\mathrm{V}^{1}$ can be calculated by Eqs. (4), (5) and (6) to determine the consistency of the distribution between them.

Our goal is to make the interlayer instance distributions close to each other. This is equivalent to making the distribution of each layer of instances close to the instances before clustering. The distributional discrepancy of instances is measured by the MIDMD and then optimized as follows:

$$\min\ D_{\mathrm{H}}^{2}\left(X,\mathrm{V}^{1}\right)=\left\|\frac{1}{n^{2}}\sum_{k=1}^{n}\sum_{k'=1}^{n}k\left(x_{k},x_{k'}\right)-\frac{2}{nc_{1}}\sum_{k=1}^{n}\sum_{i=1}^{c_{1}}k\left(x_{k},v_{i}\right)+\frac{1}{c_{1}^{2}}\sum_{i=1}^{c_{1}}\sum_{i'=1}^{c_{1}}k\left(v_{i},v_{i'}\right)\right\| \quad (7)$$

For kernel function $k(x,v)=x^{\mathrm{T}}v$, the following matrix can be obtained:

$$S_{vv}=\left((\mathrm{V}^{1})^{\mathrm{T}}\mathrm{V}^{1}\right)=\begin{pmatrix}\langle v_{1},v_{1}\rangle,...,\langle v_{1},v_{c_{1}}\rangle\\ ...\quad...\quad...\\ \langle v_{c_{1}},v_{1}\rangle,...,\langle v_{c_{1}},v_{c_{1}}\rangle\end{pmatrix} \quad (8)$$

This implies that (7) can be inferred as follows:

$$\min\ D_{\mathrm{H}}^{2}\left(X,\mathrm{V}^{1}\right)=E\left(S_{xx}\right)+E\left(S_{vv}\right)-2E\left(S_{xv}\right) \quad (9)$$

where $E(S)$ is the first moment of the matrix $S$.

## 2.3 Deep instance envelope network (MlFC&IDMD)

As the description in section 2.1, the new instance after clustering is called envelope instance. So, a deep instance envelope network-based sampling fusion method called the MlFC&IDMD is proposed by combining the proposed MlFCM and MIDMD above to achieve new instance generation. This network keeps the distribution of the instances before and after clustering consistent with each other by designing the MIDMD to avoid noisy instances. The deep instance envelope network (MlFC&MIDMD) is shown in Figure 2.

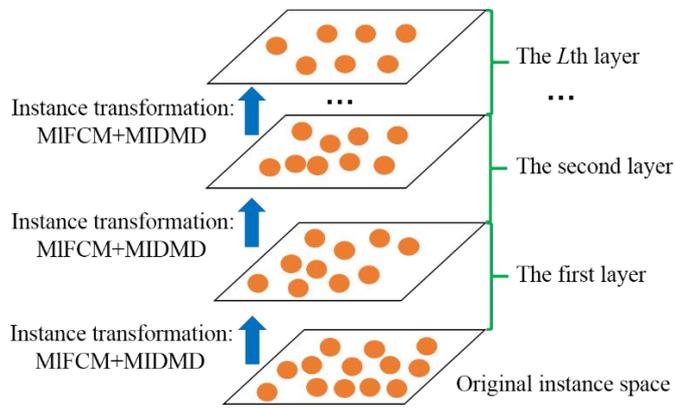

Fig. 2. *L*-layer deep instance envelope network (MlFC&IDMD)

As Figure 2 shows, first, the original instance is transformed to obtain the first layer of instances, and then the second layer of instances is obtained based on the same transformation until the *L*-layer instances are generated. The first layer instance to the *L*th layer instance constitutes the *L*-layer space (deep space). The new instances of each layer can be obtained

## 2.4 Proposed method

The number of different classes of instances is balanced by generating new instances through the proposed deep instance envelope network. Based on the MlFC&IDMD, an imbalance learning algorithm is proposed. The overall flow chart of the proposed algorithm is shown in Figure 3. For simplicity, the imbalance learning algorithm is called MlFC&IDMD&IL.

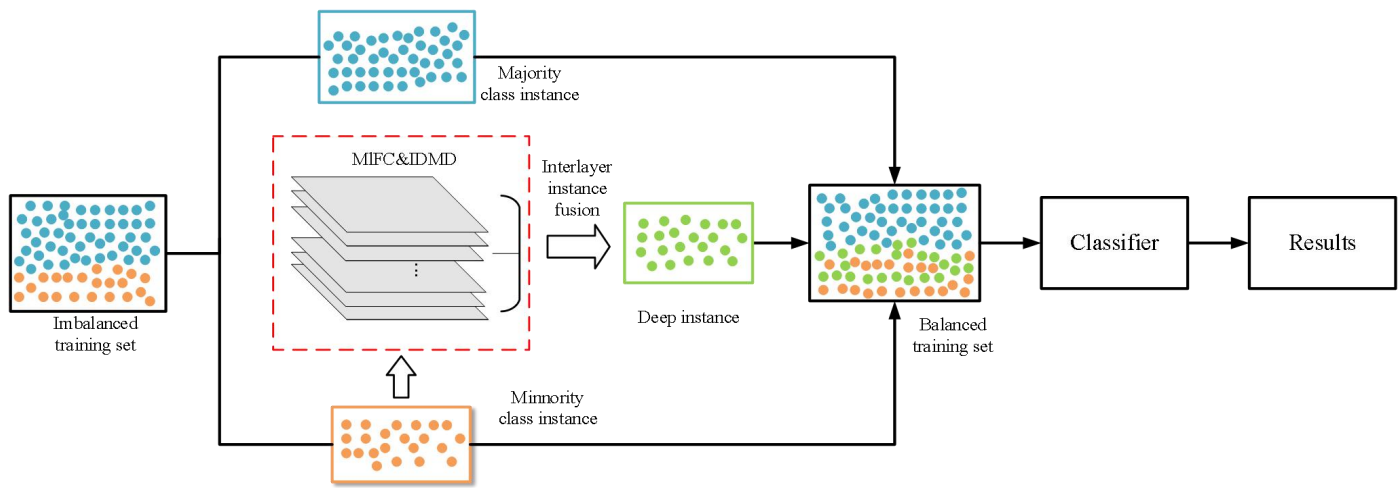

Fig. 3. Workflow of the proposed MlFC&IDMD&IL algorithm

## 3. Experimental results

### 3.1 Experimental conditions

The proposed algorithm is evaluated on 33 popular publicly available datasets. Since the proposed algorithm addresses oversampling, it is compared with the following representative oversampling methods: (1) SMOTE [18], (2) BorSMOTE [20], (3) Cluster-SMOTE (CSMOTE) [32], (4) Safety-level-SMOTE (SSMOTE) [19], (5) ADASYN [21], and (6) Random oversampling. Besides, recent algorithms, including (7) ASUWO [34], (8) K-means-SMOTE (KSMOTE) [37], (9) ADPCHFO [35], and (10) SOMO [36], are considered for comparison.

To verify the effectiveness of the MMD, the other two relevant algorithms are considered for comparison: (11) the MlFCM proposed by us, which generates deep instances without the distribution constraint term; and (12) the SMOTEMMD, which makes the distribution of the generated instances closer to that of the original instances by optimizing the MMD since SMOTE generates noise. A support vector machine (SVM) is used as the base classifier. Hold-out cross validation was adopted, and the instances were randomly divided equally into 10 groups and each group consisted of a training set and a test set at a ratio of 7:3. All experimental results discussed subsequently are computed by averaging their values across runs. The experimental platform in this section is the 64-bit Windows 7 operating system. The hardware part is a computer with an Intel(R) Core(IT) i3-4150 CPU @ 3.50 GHz processor and 8GB of RAM, and the development tool is MATLAB R2018b.

### 3.1.1 Data

Imbalanced datasets from the KEEL and UCI databases are used. Table 3 summarizes the characteristics of these datasets, including the number of instances, the number of attributes, the number of instances between categories, and the imbalance ratio (IR). These datasets are extensively employed [34-37].

Table 3. Characteristics of 33 imbalanced datasets.

| Dataset | features | samples | minority | majority | Imbalance ratio |
|---|---|---|---|---|---|
| Heart | 13 | 270 | 120 | 150 | 1.25 |
| Wine | 13 | 178 | 71 | 107 | 1.51 |
| WDBC | 30 | 569 | 212 | 357 | 1.68 |
| Wisconsin | 9 | 683 | 239 | 444 | 1.86 |
| Pima | 8 | 768 | 268 | 500 | 1.87 |
| DM | 8 | 392 | 130 | 262 | 2.01 |
| Vertebral | 6 | 310 | 100 | 210 | 2.10 |
| Yeast1 | 8 | 1484 | 429 | 1055 | 2.46 |
| Haberman | 3 | 306 | 81 | 225 | 2.78 |
| Vehicle2 | 18 | 846 | 218 | 628 | 2.88 |
| Glass | 9 | 214 | 51 | 163 | 3.19 |
| Blood-transfusion | 4 | 748 | 178 | 570 | 3.20 |
| Ecoli2 | 7 | 336 | 52 | 284 | 5.46 |
| Musk | 166 | 6598 | 1017 | 5581 | 5.48 |
| Glass6 | 9 | 214 | 29 | 185 | 6.38 |
| Yeast3 | 8 | 1484 | 163 | 1321 | 8.10 |
| Ecoli3 | 7 | 336 | 35 | 306 | 8.60 |
| Yeast-2-vs-4 | 8 | 514 | 51 | 463 | 9.08 |
| Yeast-0-3-5-9-vs-7-8 | 8 | 506 | 50 | 456 | 9.12 |
| Yeast-0-2-5-6-vs-3-7-8-9 | 8 | 1004 | 99 | 905 | 9.14 |
| Yeast-0-2-5-7-9-vs-3-6-8 | 8 | 1004 | 99 | 905 | 9.14 |
| Yeast-0-5-6-7-9-vs-4 | 8 | 528 | 51 | 477 | 9.35 |
| Ecoli4 | 7 | 336 | 20 | 316 | 15.80 |

### 3.1.2 Evaluation metrics

To assess the performance of the methods, the F-measure, AUC, and G-mean criteria were used in addition to the traditional total accuracy, and the average results on the datasets were calculated for each method. All four evaluation criteria rely on the confusion matrix shown in Table 4, where the columns and rows are the predicted classes and true classes, respectively; TP (true positive) is the number of positive class (minority class) instances correctly predicted; TN (true negative) is the number of negative class (majority class) instances correctly predicted; FP (false positive) is the number of positive negative instances incorrectly predicted; and FN (false negative) is the number of positive instances as negative incorrectly predicted.

Table 4. Confusion matrix

| | Predicted positive | Predicted negative |
|---|---|---|
| True positive | TP | FN |
| True negative | FP | TN |

The total accuracy (Acc) can be used to test the classification performance of a model once the dataset has reached a balanced state. For a balanced dataset in which two classes are identically represented, the total accuracy is a prevalent choice to measure classification performance. It is calculated as follows:

$$\text{Acc} = \frac{TP + TN}{TP + FP + FN + TN} \tag{24}$$

The F-measure is one of the most commonly used criteria [46] in the field of imbalanced datasets. To calculate the F-measure, the precision and recall must be calculated first as follows:

$$\text{Precision} = \frac{TP}{TP + FP} \tag{25}$$

$$\text{Recall} = \frac{TP}{TP + FN} \tag{26}$$

The precision and recall are two important performance measures for minority class instances, and the F-measure combines these two metrics into a single average. In principle, the F-measure represents the harmonic mean of the recall and precision. Therefore, a higher F-measure ensures that both the recall and precision are quite high, and a higher

F-measure means that the desired method correctly classifies minority class instances at a higher rate and with fewer misclassifications, i.e., the F-measure evaluates the accuracy of the classifier on only the minority classes [47-48]. The calculation formula for the F-measure is as follows:

$$\text{F-measure} = 2 * \frac{\text{precision}*\text{recall}}{\text{precision}+\text{recall}} \tag{27}$$

The AUC is another criterion commonly used for imbalanced data problems [49]. For a classifier, the AUC is the area under the corresponding ROC graph; and the larger the area is, the better the performance of the classifier. This criterion is calculated as follows:

$$\text{AUC} = \frac{Sensitivity + Specificity}{2} \tag{28}$$

where sensitivity and specificity are calculated as follows.

$$Sensitivity = \frac{TP}{TP + FN} \tag{29}$$

$$Specificity = \frac{TN}{TN + FP} \tag{30}$$

The G-mean is the geometric mean of the true positive rate and the true negative rate, and it is calculated as follows:

$$\text{G-mean} = \sqrt{\frac{TP}{TP+FN} \cdot \frac{TN}{TN+FP}} \tag{31}$$

### 3.1.3 Nonparametric statistical tests for statistical analysis

A nonparametric statistical test known as the Friedman test [50] was used to assess the statistical performance of the algorithm. According to this test, first, competing algorithms are assumed to be equivalent in terms of performance (null hypothesis). The main objective is to determine whether the null hypothesis will be rejected. Rejection of this hypothesis implies that there is a significant difference between algorithms. This difference is determined by calculating the p-value; and if the p-value is below or equal to the defined significance level, the null hypothesis will be rejected. In this study, $\alpha = 0.05$ was considered the level of significance, and a post hoc procedure such as Holm's test [51] can be applied to determine the differences between methods. In our experiments, the p-values were obtained by Holm's test.

### 3.2 Algorithm comparison

To verify the quality of the instances generated by the MlFC&IDMD network, two groups of experiments were organized: 1) Compare the results of average F-measure (F-M), AUC and G-mean (G-M) before and after balancing the training set, and the 'none' method represents the imbalanced training set is applied without resampling; 2) The intra-class variance of the original minority instances and the inter-class variance with the majority instances were calculated separately. For the balanced minority instances, we twice randomly selected instances equal to the original minority instances size and calculated their intra-class variance and inter-class variance with the majority instances. Six datasets are chosen, including DM, Blood-transfusion, Glass6, Yeast-0-2-5-7-9-vs-3-6-8, Ecoli4 and Yeast5; and they represent two types of high- and low-IR datasets. The results are recorded in Tables 5-6.

Table 5. Comparison of classification results before and after balancing training set

| Dataset | DM | | | | Blood-transfusion | | | |
|---|---|---|---|---|---|---|---|---|
| Measure | Acc | AUC | F-M | G-M | Acc | AUC | F-M | G-M |
| MlFC&IDMD&IL | **0.7805±0.0349** | **0.7751±0.0475** | **0.6940±0.0571** | **0.7737±0.0481** | 0.6431±0.0395 | **0.6467±0.0282** | **0.4648±0.0323** | **0.6385±0.0280** |
| None | 0.7712±0.0397 | 0.7116±0.0413 | 0.6082±0.0627 | 0.6891±0.0448 | **0.7600±0.0000** | 0.5000±0.0000 | 0.0000±0.0000 | 0.0000±0.0000 |
| Dataset | Glass6 | | | | Yeast-0-2-5-7-9-vs-3-6-8 | | | |
| Measure | Acc | AUC | F-M | G-M | Acc | AUC | F-M | G-M |
| MlFC&IDMD&IL | **0.9631±0.0149** | **0.9413±0.0433** | **0.8717±0.0549** | **0.9397±0.0451** | 0.9513±0.0123 | **0.8974±0.0353** | **0.7725±0.0539** | **0.8941±0.0376** |
| None | 0.9569±0.0159 | 0.9190±0.0501 | 0.8464±0.0604 | 0.9161±0.0526 | **0.9613±0.0098** | 0.8258±0.0511 | 0.7671±0.0700 | 0.8060±0.0623 |
| Dataset | Ecoli4 | | | | Yeast5 | | | |
| Measure | Acc | AUC | F-M | G-M | Acc | AUC | F-M | G-M |
| MlFC&IDMD&IL | **0.9861±0.0083** | **0.9302±0.0521** | **0.8809±0.0697** | **0.9264±0.0564** | **0.9707±0.0081** | **0.9054±0.0480** | **0.6441±0.0774** | **0.9013±0.0525** |
| None | 0.9634±0.0067 | 0.6917±0.0562 | 0.5452±0.1232 | 0.6123±0.0964 | 0.9687±0.0000 | 0.5000±0.0000 | 0.0000±0.0000 | 0.0000±0.0000 |

Table 6. Comparison of the performance of minority class instances before and after balancing

| Dataset | DM | | | | Blood-transfusion | | | |
|---|---|---|---|---|---|---|---|---|
| Measure | Intra-class variance | | Inter-class variance | | Intra-class variance | | Inter-class variance | |
| MlFC&IDMD | **0.1326** | **0.1600** | **0.0350** | **0.0323** | **0.0721** | **0.0766** | **0.0028** | **0.0026** |
| Original | 0.2261 | 0.2261 | 0.0271 | 0.0271 | 0.0981 | 0.0981 | 0.0023 | 0.0023 |
| Dataset | Glass6 | | | | Yeast-0-2-5-7-9-vs-3-6-8 | | | |
| Measure | Intra-class variance | | Inter-class variance | | Intra-class variance | | Inter-class variance | |
| MlFC&IDMD | **0.2051** | **0.2045** | **0.0797** | **0.0834** | **0.0957** | **0.1012** | **0.0159** | **0.0228** |
| Original | 0.2427 | 0.2427 | 0.0719 | 0.0719 | 0.1167 | 0.1167 | 0.0134 | 0.0134 |
| Dataset | Ecoli4 | | | | Yeast5 | | | |
| Measure | Intra-class variance | | Inter-class variance | | Intra-class variance | | Inter-class variance | |
| MlFC&IDMD | **0.2138** | **0.2124** | **0.0240** | **0.0234** | **0.1050** | **0.1078** | **0.0154** | **0.0145** |
| Original | 0.2195 | 0.2195 | 0.0143 | 0.0143 | 0.1090 | 0.1090 | 0.0066 | 0.0066 |

From Table 5, it can be seen the performance of the proposed algorithm is greatly improved on all four metrics compared to the imbalanced dataset (see the boldface type). In particular, for Blood and Yeast5, since the trained model for the imbalanced dataset is biased towards the majority class, all minority classes in the test set are predicted to be majority classes and the values of AUC, F-M and G-M are 0.5,0,0. After the datasets reach balanced, the performance of all three metrics has improved significantly. This indicates the MlFC&IDMD network generates high quality instances and is effective.

As can be seen from Table 6, compared with the original minority instances, the intra-class variance decreases and the inter-class variance becomes better for the same number of balanced minority instances, and there is also a difference between the two random selection (see the boldface type). In this table, each column represents the result of one selection, for example, for DM, 0.1326 represents the intra-class variance calculated from balanced minority instances after the first selection, 0.1600 represents the intra-class variance of the second selection, 0.0350 represents the inter-class variance of the first selection, and 0.0323 represents the inter-class variance of the second selection. Because the original minority instances have been kept constant, the intra-class and inter-class variances remain the same for the two selection cases. The results indicate that the MlFC&IDMD network generate instances with more diversity and better quality.

Tables 5-6 clarify that the new instances generated by the MlFC&IDMD network make the quality of the minority class instances improved. Table 7 also adopts the average Acc, F-M, AUC and G-M for the evaluation criteria, and the best results for each dataset are shown in bold. The results for the two datasets are shown in this section, and the complete results of Table 7 are placed in the Appendix section.

Table 7. Results of the oversampling methods on all test datasets

| Dataset | Heart | | | | Wine | | | |
|---|---|---|---|---|---|---|---|---|
| Measure | Acc | AUC | F-M | G-M | Acc | AUC | F-M | G-M |
| MlFC&IDMD&IL | **0.8160±0.0326** | **0.8128±0.0338** | **0.7906±0.0399** | **0.8115±0.0343** | 0.9564±0.0299 | 0.9523±0.0329 | 0.9443±0.0386 | 0.9517±0.0336 |
| MlFCM | 0.8136±0.0316 | 0.8103±0.0323 | 0.7879±0.0379 | 0.8090±0.0328 | 0.9545±0.0335 | 0.9515±0.0347 | 0.9429±0.0413 | 0.9511±0.0348 |
| SMOTE | 0.8123±0.0301 | 0.8097±0.0318 | 0.7876±0.0383 | 0.8083±0.0323 | 0.9509±0.0298 | 0.9462±0.0314 | 0.9376±0.0373 | 0.9456±0.0317 |
| SMOTEMMD | 0.7938±0.0335 | 0.7953±0.0327 | 0.7770±0.0355 | 0.7940±0.0336 | **0.9655±0.0263** | **0.9629±0.0280** | **0.9564±0.0331** | **0.9627±0.0281** |
| BorSMOTE | 0.8049±0.0353 | 0.8050±0.0360 | 0.7855±0.0413 | 0.8040±0.0364 | 0.9527±0.0345 | 0.9492±0.0354 | 0.9405±0.0425 | 0.9487±0.0356 |
| CSMOTE | 0.8111±0.0267 | 0.8086±0.0288 | 0.7864±0.0351 | 0.8073±0.0294 | 0.9455±0.0284 | 0.9409±0.0285 | 0.9311±0.0345 | 0.9403±0.0287 |
| SSMOTE | 0.8012±0.0267 | 0.7989±0.0355 | 0.7768±0.0390 | 0.7973±0.0366 | 0.9545±0.0261 | 0.9508±0.0277 | 0.9424±0.0328 | 0.9503±0.0281 |
| ASUWO | 0.8012±0.0506 | 0.7994±0.0505 | 0.7781±0.0551 | 0.7985±0.0500 | 0.9400±0.0285 | 0.9311±0.0330 | 0.9213±0.0388 | 0.9294±0.0346 |
| ADASYN | 0.7988±0.0428 | 0.7978±0.0425 | 0.7770±0.0468 | 0.7967±0.0428 | 0.9455±0.0297 | 0.9417±0.0295 | 0.9315±0.0358 | 0.9410±0.0296 |
| Random | **0.8160±0.0326** | 0.8125±0.0334 | 0.7901±0.0392 | 0.8112±0.0339 | 0.9473±0.0249 | 0.9417±0.0255 | 0.9328±0.0309 | 0.9409±0.0256 |
| KSMOTE | 0.7333±0.0382 | 0.7417±0.0321 | 0.7306±0.0296 | 0.7317±0.0397 | 0.9460±0.0242 | 0.9400±0.0254 | 0.9306±0.0315 | 0.9393±0.0255 |
| ADPCHFO | 0.8111±0.0329 | 0.8086±0.0332 | 0.7869±0.0385 | 0.8073±0.0338 | 0.9455±0.0227 | 0.9386±0.0243 | 0.9298±0.0295 | 0.9376±0.0250 |
| SOMO | 0.8037±0.0316 | 0.8003±0.0319 | 0.7767±0.0375 | 0.7985±0.0321 | 0.9509±0.0298 | 0.9462±0.0314 | 0.9376±0.0373 | 0.9456±0.0317 |

As seen from Table 7, the proposed MlFC&IDMD&IL method generally works best as it obtains the optimal results in at least one performance measure on 33 datasets the most at 54 times. The SMOTEMMD is second with 15 times; and SOMO, ASUWO, and KSMOTE follow closely with the best results 12, 9, and 8 times, respectively. Specifically, the MlFC&IDMD&IL has higher Acc values for 14 datasets, has higher AUC values for 10 datasets, has higher F-measures for 20 datasets and has higher G-M values for 10 datasets compared to other oversampling methods. Thus, the results further show that the proposed method performs best on all four criteria, especially F-M. The results also show that the proposed algorithm works better on datasets with a higher IR such as the Yeast-1-2-8-9-vs-7, Poker-9-vs-7 and Yeast4 datasets compared to other datasets. The possible reason is that the larger IR is, more layers in the network space lead to the generation of more informative instances.

To evaluate the classification performance of different methods on multiple datasets, the mean rankings of the Acc, F-measure, AUC and G-mean of each method on these datasets were derived and analyzed. The method with the best performance was ranked first while the worst performing method was ranked thirteenth. Thus, the best method has the lowest mean ranking. The mean rankings of each method on the 33 datasets are obtained and recorded in Figure 4.

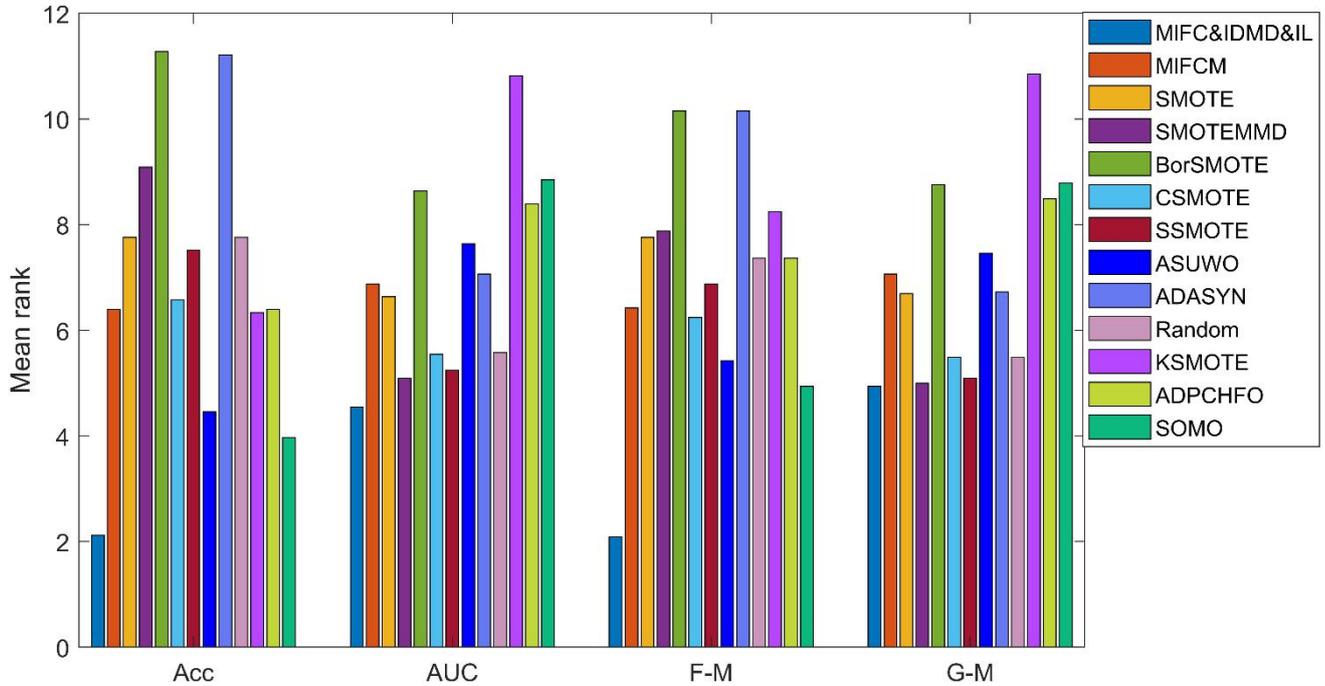

Fig. 4. Mean ranking of all compared oversampling techniques on all tested datasets for the SVM

It is also obvious from Figure 4 that the rankings of other methods differs significantly from those of the MlFC&IDMD&IL on the Acc and F-M, indicating that the deep instances generated by the MlFC&IDMD are more advantageous, and also the method performs optimally on the AUC and G-M. Additionally, the SMOTEMMD is ranked 5.091 and 5.000 on the AUC and G-M metrics, respectively, which is better than SMOTE's rankings of 6.636 and 6.697, respectively. Comparing all algorithms, the SMOTEMMD ranks second on the AUC and G-M. Similarly, compared to the MlFCM, the MlFC&IDMD&IL outperforms the MlFCM on all four metrics. Therefore, distributional consistency can facilitate the generation of more useful instances, reduce noisy instances, and improve the classification accuracy of instances.

## 4. Discussion and conclusion

Imbalanced learning is important and challenging since the class imbalance problem poses a difficult task in machine learning and data mining fields. Existing approaches aimed at solving the class imbalance problem can be categorized as data level, algorithm level, ensemble learning and feature selection approaches. Resampling as a data-level approach is an effective way to solve this problem independently of classifier selection. However, random oversampling often leads to overfitting, which in turn reduces the performance of the model. Although SMOTE-based oversampling can avoid overfitting, they tend to generate noisy instances easily; moreover, most of the oversampling methods, such as the safe-level-SMOTE, Borderline-SMOTE and ADASYN, aim to solve the between-class imbalance problem while ignoring the within-class imbalance problem. Therefore, clustering-based oversampling methods are proposed to solve the within-class imbalance problem. However, current clustering-based oversampling methods are mainly single-level clustering, which ignores the diversity of instance information. Single-level clustering has difficulty mining more instance information and heavily relies on prior knowledge. Although multilayer clustering algorithms have been proposed such as HGC algorithm, its purpose is still clustering rather than new instances generation.

To solve the problem, an oversampling method called the MlFC&IDMD&IL that combines the proposed MlFCM and MIDMD is proposed. The MlFCM is designed to solve the problem of single-layer clustering since multilayer clustering can mine more instance information to obtain high-quality instances. The MIDMD is designed to solve the problem of noisy instances by reducing the instance distribution discrepancy to avoid the generation of noisy instances. The main flow of the proposed algorithm is as follows: First, the MlFCM is designed to obtain deep instances from the original minority class instances and increase the diversity of instances. Then, a minimum interlayer discrepancy mechanism is proposed to avoid the generation of noisy instances. Next, the multilayer FCM and minimum interlayer discrepancy mechanism are combined to construct a deep instance envelope network – the MlFC&IDMD. Finally, an imbalanced learning algorithm is proposed based on the MlFC&IDMD. Thirty-three public imbalanced datasets are used for verification. The SVM classifier is used, and over ten representative relevant algorithms are used for comparison. The experimental results show that the new instances generated by the MlFC&IDMD network make the quality of the minority class instances improved and the proposed algorithm significantly outperforms other popular methods on the four metrics of the Acc, AUC, F-M and G-M. It is worth mentioning that the MlFC&IDMD&IL performs better than the KSMOTE, ADPCHFO, ASUWO, and SOMO clustering oversampling methods on all four performance criteria. This indicates that multilayer clustering is more beneficial for mining more instance information than single-layer clustering, so multilayer clustering is feasible and is more favorable for enriching instance information. The MlFC&IDMD&IL has more advantages over the MlFCM and SMOTE, which means that consistency of the instance distribution can help to avoid the generation of noisy instances and improve the classification performance of minority class instances.

The main contributions of this paper are as follows.

1) For the first time, a multilayer FCM (MlFCM) algorithm is proposed to mine the instances for more information, and the deep instances are fused with the original instances, which can enrich the information of the original instances, making the minority instances more diverse and more conducive to classification.

2) A minimum interlayer discrepancy mechanism (MIDMD) is proposed to make the distribution of the instances before and after clustering consistent with each other, avoiding the problem of generating noise due to the inconsistent distribution of instances generated by methods such as the SMOTE.

3) A deep instance envelope network (MlFC&IDMD) is constructed by combining the MlFCM and MIDMD, thereby generating high quality deep instances.

4) An efficient strategy is proposed to determine the number of layers $L$ of the deep instance envelope network, and a relationship between L and the degree of imbalance ratio IR is established in this paper.

5) A new imbalance learning algorithm is proposed based on the MlFC&IDMD.

The proposed algorithm provides a new way to solve the class imbalance problem, and it is not limited to specific clustering algorithms and instance distribution metric algorithms; therefore, it can be extended to any clustering algorithm and instance distribution metric algorithm. Since the proposed oversampling method can be used to balance any dataset and is independent of the classifier, the method has good generalization capability and can also be applied to real-world class imbalance problems such as anomaly detection, medical diagnosis, target identification, and business analytics.

Although this paper demonstrates the effectiveness of the proposed method, future work remains. First, the main weakness of the proposed method is the increased complexity. The possible reason may be related to the number of layers of the deep instance envelope network and the number of clusters per layer. Therefore, future work is to optimize the entire network space. Second, it is worth noting that more multiclass imbalanced classification problems will be considered in future work.

## Acknowledgments

The authors would like to thank the editor and reviewers for their valuable comments and suggestions. The authors would also like to thank those individuals or institutions that have provided data support for this research. This work was supported in part by the National Natural Science Foundation of China (NSFC) under grant 61771080, the Key Project of Technology Innovation and Application Development in Chongqing (cstc2019jsxz-mbdxX0050),the    Natural Science Foundation of Chongqing (cstc2020jscx-msxm0369, cstc2020jcyj-msxmX0100, cstc2020jscx-fyzx0212, and grant 2020ZY023659), the Basic and Advanced Research Project in Chongqing (cstc2020jcyj-msxmX0523, cstc2020jscx-fyzx0212, and grant 2020ZY023659), and The Chongqing Social Science Planning Project (2018YBYY133).

## Declarations of interest

None.

# Consent for publication

Not applicable

# Data availability

The codes and data can be found in:

The data and code is in (cloud storage): https://pan.baidu.com/s/1ptQgNUZrNWmcw264Py46SA , extraction code: lbdv

The data and code is in (Github): https://github.com/leaphan/imbalance-learning .